\documentclass[11pt,a4paper]{article}
\usepackage[hyperref]{acl2020}
\usepackage{times}
\usepackage{amsmath}
\usepackage{graphics}
\usepackage{latexsym}
\usepackage{arydshln}
\usepackage{graphicx}
\usepackage{footnote}
\usepackage{algpseudocode}%
\usepackage[linesnumbered,ruled]{algorithm2e}

\usepackage{microtype}

\aclfinalcopy 
\def\aclpaperid{2463} 

\title{Exploring Neural Models for Parsing Natural Language into \\First-Order Logic}

\author{Hrituraj Singh \thanks{This work was done when the author was working as a Software Developer at Adobe Systems} \\
  Adobe Research\\\And
  Milan Aggrawal \\
  Adobe Systems  \\\And
  Balaji Krishnamurthy \\
  Adobe Systems \\}
\date{}

\begin{document}
\maketitle
\begin{abstract}
Semantic parsing is the task of obtaining machine-interpretable representations from natural language text. We consider one such formal representation - First-Order Logic (FOL) and explore the capability of neural models in parsing English sentences to FOL. We model FOL parsing as a sequence to sequence mapping task where given a natural language sentence, it is encoded into an intermediate representation using an LSTM followed by a decoder which sequentially generates the predicates in the corresponding FOL formula. We improve the standard encoder-decoder model by introducing a variable alignment mechanism that enables it to align variables across predicates in the predicted FOL. We further show the effectiveness of predicting the \textit{category} of FOL entity - Unary, Binary, Variables and Scoped Entities, at each decoder step as an auxiliary task on improving the consistency of generated FOL. We perform rigorous evaluations and extensive ablations. We also aim to release our code as well as large scale FOL dataset along with models to aid further research in logic-based parsing and inference in NLP.
\end{abstract}

\section{Introduction}
Semantic parsing aims at mapping natural language to structured meaning representations. This enables a machine to understand unstructured text better which is central to many tasks requiring natural language understanding such as question answering \cite{berant2013semantic,pasupat2015compositional}, robot navigation \cite{macmahon2006walk,artzi2013weakly}, database querying \cite{zelle1996learning} etc. For question answering, natural language question is converted to formal semantics which facilitates interaction with a knowledge base (such as FreeBase \cite{bollacker2008freebase}) for retrieving concise answers \cite{furbach2010logic}. Such representations can be used to specify instructions to robots \cite{artzi2013weakly} or conversational agents \cite{artzi2011bootstrapping} for executing desired action(s) in an environment. Similarly, natural language queries are transformed into executable database programming language instructions (such as SQL) to retrieve or generate correct results in a database \cite{sun_etal_2018_semantic,zhong2017seq2sql}.

A variety of logical forms and meaning representations have been proposed for text. These include graph-based formalisms \cite{banarescu2013abstract,abend2013universal,oepen2014semeval,kollar2018alexa} where text is represented as a typed graph. The entities and action events are represented as nodes with labeled edges depicting relations between them. Semantic dependency tree \cite{oepen2014semeval} is a directed graph depicting the syntactic structure of a sentence in the form of modifier relations between its words. AMR (Abstract Meaning Representation) graphs \cite{banarescu2013abstract} use variables to annotate nodes following neo-Davidsonian style \cite{davidson1969individuation}. Lambda Dependency-based Compositional Semantics ($\lambda$-DCS) \cite{liang2013lambda} was proposed as a formal language adapting Dependency-Based Compositional Semantics \cite{liang2013learning} borrowing the expressiveness of lambda calculus \cite{barendregt1984lambda} but aiming to remove explicit use of variables.

In this work, we focus on first-order logic (FOL) \cite{smullyan2012first} as the language formalism for text. FOL represents entities and actions in natural language through quantified variables and consists of functions (called predicates) which take variables as arguments. The predicates attach semantics to variables and express relations between objects \cite{blackburn2005representation}. For instance, a simple sentence - ``a man is eating" can be represented through FOL as

\begin{center}
    $\exists A (\exists B (man(A) \ \land \ eat(B) \  \land \ agent(B,A)))$ 
\end{center}


Advanced natural language concepts as in sentence ``the man and woman are seated facing each other" can be expressed as

\begin{center}
    $\exists A (\exists B(\exists D (\exists C(man(A)\ \land \ woman(B) \ \land \ seat(D) \ \land \  \textcolor{blue}{subset\_of(A,C)} \ \land \ \textcolor{blue}{subset\_of(B,C)} \ \land \ theme(D,C) \ \land \ \textcolor{red}{not(\exists E(other(E)} \ \land \ \textcolor{orange}{not(\exists}  \textcolor{orange}{F(face(F)} \ \land \ \textcolor{orange}{theme(F,E)} \ \land \ \textcolor{orange}{agent(F,C)))} \ \textcolor{red}{))}))))$
\end{center}

where ``man" and ``woman" are \textcolor{blue}{represented together} through shared variable \textcolor{blue}{C} and ``facing each other" is represented by \textcolor{red}{negating} the existence of a thing $\textcolor{red}{E}$ for which \textcolor{orange}{C is not facing E} holds true.


The success of learning based neural approaches in NLP tasks like machine translation \cite{cho2014learning,sutskever2014sequence,vaswani2017attention}, paraphrase generation \cite{prakash2016neural,gupta2018deep}, dialog modeling \cite{vinyals2015neural,kottur2017exploring}, machine comprehension \cite{wang2017conditional}, logical inference \cite{kim2019semantic} has motivated their use for semantic parsing \cite{kovcisky2016semantic,buys2017robust,cheng2017learning,liu2018discourse,li2018seq2seq} as well. Many such works use the encoder-decoder framework to model it as a sequence transduction task. Since they were designed for solving specific tasks like question answering, such methods \cite{jia2016data,dong2016language} have mainly focussed on confined logical formalism for specific domains such as flight reservation, restaurant booking, etc \cite{wang2015building} capturing limited vocabulary and semantic concepts.


In this paper, we aim at developing a general-purpose open-domain neural first-order logic parser for natural language sentences to examine the capabilities of such models. We train our model by obtaining a large corpus of text-FOL pairs for sentences in SNLI Dataset \cite{bowman2015large} through C\&C parser \cite{clark2007wide} and Boxer \cite{bos2008wide} (discussed later in detail).\footnote{https://github.com/valeriobasile/candcapi} Apart from meaning depiction, parsing sentences to FOL would enable neural models to capture complex relationships between entities resulting in richer embeddings which might be useful in several other NLP tasks. Such an examination would help understand challenges in generating FOL through neural approaches owing to complexities in its representation. Since it is one of the first such exploration for FOL, we treat the popular sequence to sequence model coupled with attention mechanism \cite{bahdanau2014neural} as our baseline. We propose to disentangle the prediction of different types of FOL syntactic entities (unary and binary predicates, variables etc) while parsing sentences and show improvements through performing category type prediction as an auxiliary task. We further show major improvements by explicitly constraining the decoder to align variables across unary and binary predicates. This restricts the model to maintain consistency while expressing standalone entity attributes and relations between them. 

Our contributions can be enumerated as: 1) We explore and develop an open domain neural semantic parser to parse natural language sentences to FOL using Seq2Seq framework; 2) We propose disentangled FOL entity type prediction along with FOL parsing under multi-task learning and FOL variable alignment through decoder alignment mechanism. We perform extensive ablation studies to establish the improvements registered; 3) We also aim to release our code, models and large scale dataset used comprising of sentence-FOL mappings to aid further research in FOL based NLP.



\section{Background}
\label{section:background}

\textbf{Text to FOL Conversion : }In this section, we give a brief overview of syntactic-semantic analysis pipeline used for obtaining the mappings data through Boxer \cite{bos2008wide} based on Combinatory Categorial Grammar (CCG) \cite{steedman2011combinatory} and  Discourse Representation Theory (DRT) \cite{kamp2011discourse}. CCG is phrase-level grammar which defines rules for generating constituency-based structures. CCG comprises of syntactically typed lexical items such that each item is a lambda-expression and uses combinatorial logic (lambda calculus) to combine them through the application of combinators. CCG derivation guides semantic composition to obtain Discourse Representation Structures (DRS) from CCG parses. DRS comprises of discourse referents and conditions defined on them which can be recursive. DRS is capable of representing varied linguistic phenomena such as anaphora, presupposition, tense and aspect. These DRSs are compatible and can be converted to FOL through a set of syntactic transformations \cite{bunt2001patrick}. Formally, predicates in FOL are atomic formulas that are combined through logical connectives - logical and ($\land$), logical or ($\lor$) ; and quantifiers. In general, a predicate $P(v_1,v_2, ... , v_n)$ is an n-ary function of variables. There are two types of quantifiers, universal ($\forall$) - which specifies that sub-formula within its scope is true for all instances of the variable and existential ($\exists$) - which asserts existence of at least one instance represented by a variable under which the sub-formula holds true. For example, ``All humans eat" can be represented as 
\begin{center}
    $\forall A (\exists B (human(A) \ \land \ eat(B) \ \land \ agent(B,A)))$
\end{center}

Following generalized De Morgan's law \cite{johnstone1979conditions}, universal quantifiers can be represented through existential quantification and negation ($not$) preserving the semantics as

\begin{center}
    $not(\exists A(not(\exists B(human(A) \ \land \ eat(B) \ \land \ agent(B,A)))))$
\end{center}

\vspace{2 mm}
\noindent \textbf{Output and Mapping Format : }Given a text sentence, we obtain the following FOL output.

\vspace{2 mm}
\noindent \textit{\textbf{Sentence :} }``three women are traveling by foot" 

\vspace{1 mm}
\noindent \textit{\textbf{Output FOL : } } fol(1,some(A,some(B,some(C,and (r1by(B,A),and(n1foot(A),and(r1agent(B,C),and (v1travel(B),and(n1woman(C),some(D,and(card (C,D),and(c3number(D),n1numeral(D)))))))))))))

\vspace{1 mm}
\noindent Here, the predicates are prefixed with POS-tags \cite{wilks1998grammar} and relation types. Since the output FOL comprises of existential quantifiers and disjunction of atomic formulas only, we convert it into an equivalent mapping as a sequence of predicates, argument variables, scoping symbols (such as ``fol(", ``)", ``not(") and train our models to predict the sequence. We arrange scope symbols in accordance to their nesting level (top most appearing first in the sequence) with further ordering that entities that are part of same scope are arranged as sequence of unary predicates, followed by binary predicates and other nested scoped entities.

\vspace{1 mm}
\noindent \textit{\textbf{Equivalent Mapping : }}fol( n1foot A v1travel B n1woman C c3number D n1numeral D r1by B A r1agent B C card C D )

\section{Proposed Approach}
We model parsing a given sentence into FOL as a sequence to sequence \textit{transduction} problem. Our parser $P$ generates a token in the output FOL representation in a sequential manner by greedily sampling it from a probability distribution conditioned on the input sentence and the previously generated tokens. Our input $X$ consists of a sequence of $m$ tokens $\{x_0, x_1,...,x_m\}$ which get encoded into hidden \textit{contextual} representations by an Encoder. The Decoder, then, generates an output sequence of $n$ tokens $\{y_0, y_1,...,y_n\}$.
\begin{equation}
    P(Y|X) = \prod_{i=0}^{n} P(y_i|y_0..y_{i-1},X)
\end{equation}

\subsection{Encoder}
Our Encoder $E$ is a bidirectional LSTM (biLSTM) which encodes a sequence of input tokens $X: \{x_0,x_1...,x_m\}$ into a sequence of hidden states $H_e : \{h_{e_0},h_{e_1},...,h_{e_m}\}$, $h_{e_i} \in {R}^{d_h}$ to capture contextual information from the input that is eventually used by the decoder to produce the output FOL sequence. The biLSTM block takes word embeddings for the input tokens $E_e: \{e_{e_0},e_{e_1},...,e_{e_m}\}$, $e_{e_i} \in R^D$  as input and processes them to calculate the contextual representations
\begin{equation}
   h_{e_i} = [h_{fe_i};h_{be_i}] = biLSTM(e_{e_0}, e_{e_1},...,e_{e_i})
\end{equation}
where $;$ denotes concatenation operation and $h_{fe_i}, h_{be_i}$ refer to the forward and backward hidden states of the biLSTM.

\subsection{Decoder}\label{decoder}
Decoder $D$ consists of an LSTM which uses the outputs of encoder $E$ along with previously decoded outputs, provided as embeddings $E_d : \{e_{d_0},e_{d_1},...,e_{d_n}\} $ to it as input, to generate a sequence of hidden states $H_d : \{h_{d_0},h_{d_1},...,h_{d_n}\}$. 
\begin{equation}
    h_{d_i} = LSTM(e_{d_0}, e_{d_1},...,e_{d_i})
\end{equation}
Attention \cite{bahdanau2014neural} has now become ubiquitous in sequence to sequence models. We consider it to be a part of our baseline model. Following \cite{bahdanau2014neural}, we calculate the weights for encoder-decoder attention using $H_d$ as \textit{queries} while $H_e$ as \textit{keys} as well as \textit{values} (eq. \ref{enc-dec-attn}.
\begin{equation}
    \alpha_{ij} = \frac{e^{h_{d_i}^Th_{e_j}}}{\sum_{k=0}^{m}e^{h_{d_i}^T h_{e_k}}}
\label{enc-dec-attn}
\end{equation}

\begin{figure*}[t]
\centering
\includegraphics[width=0.95\textwidth,height=2.5in]{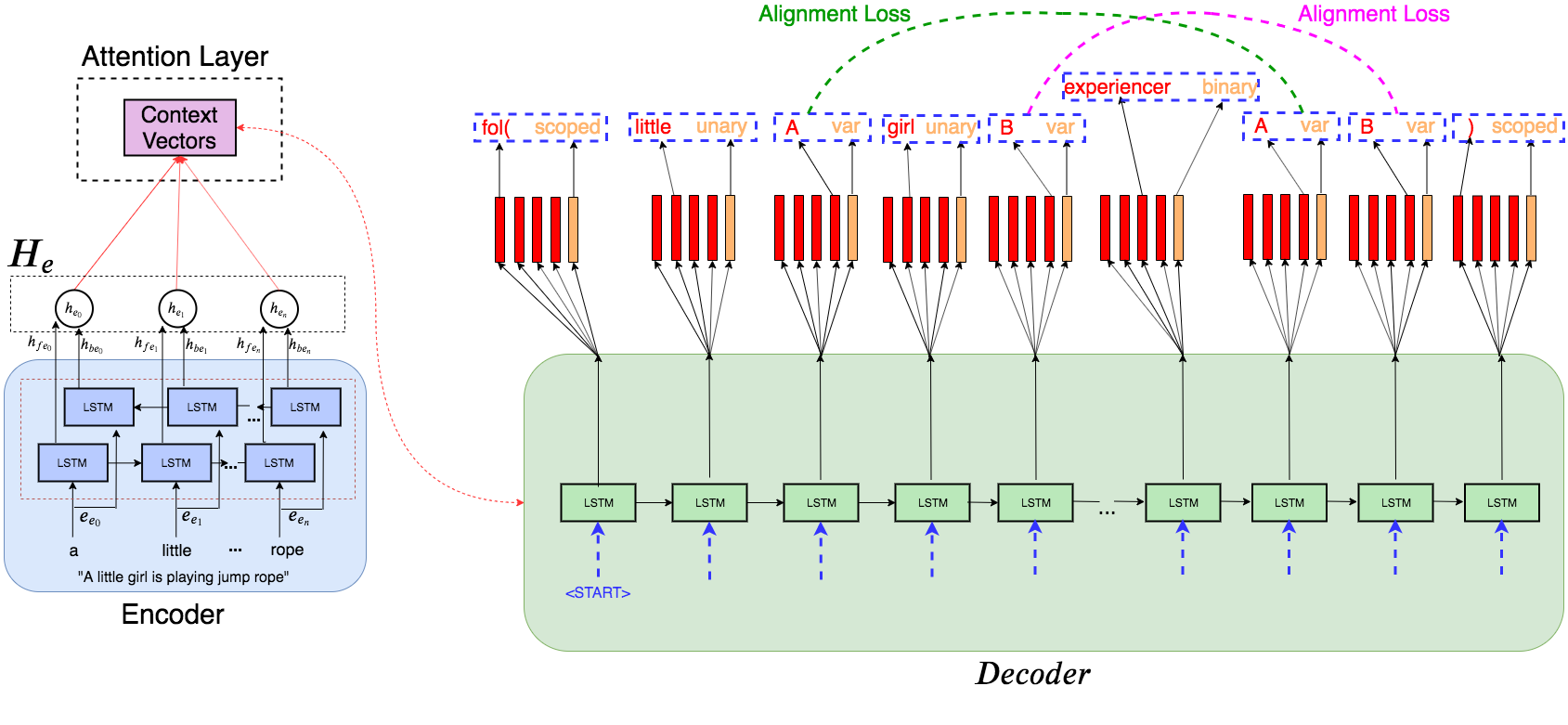}

\caption{Overview of our architecture showing separate heads (red), category prediction (orange) and alignment mechanism (green and pink). Input to Decoder LSTM (blue) depicts the output of last step being fed at next step. Red arrow between Attention Layer and Decoder depicts standard encoder-decoder attention.}
\label{fig:arch}
\end{figure*} 

The encoder-context vector is obtained by taking a weighted sum of encoder's hidden states $H_e$ (eq. \ref{enc-dec-context}).
\begin{equation}
    c_{e_i} = \sum_{j=0}^{m}\alpha_{ij} * h_{e_j}
\label{enc-dec-context}
\end{equation}

The hidden state of the decoder $h_{d_i}$ along with encoder-context vector $c_{e_i}$ is used to predict the final output token at $i^{th}$ step.
\begin{equation}
    s_i = W_o * [h_{d_i};c_{e_i}]
\end{equation}
where $W_o \in R^{V X (d_h+d_c)}$ is output \textit{head} and $d_h$ is the dimension of hidden vector and $d_c = d_h$ is the dimension of context vector.

We train the model on the standard cross-entropy objective while adopting \textit{teacher forcing} methodology \textit{i.e.} giving the inputs to the decoder from ground truth instead of previously decoded tokens while training
\begin{equation}
    L_{CE} (\theta) = -\frac{1}{n}\sum_{0}^n{y_{i}^t* log(P(y_i|y_0..y_{i-1},X))}
\end{equation}
where $y_i^t$ is the target token from ground truth at step $i$ and $\theta$ refers to the trainable model parameters.

\subsubsection{Separate Heads}\label{sep-heads}
The output tokens in an FOL sequence do not all belong to the same token category unlike majority sequence to sequence translation problems which process words. In particular, the output tokens in an FOL sequence can be divided into four major types - Unary Predicates $U$, Binary Predicates $B$, Variables $V$, and Scoped Entities $S$. We create separate vocabularies of sizes $V_u$, $V_b$, $V_v$, and $V_s$ for each category. Apart from variables $V$ which have one-hot embedding, all other types of output tokens have \textit{dense} embeddings. This is because a token of category $V$ does not posses semantic meaning that is shared across all sequences from the output distribution. Thus, they are defined in the context of an FOL sequence only. We represent them through one-hot embeddings to ensure independence between them.

Building on above motivation, we use five different \textit{heads} on top of Decoder LSTM. While one head $T$ decides what type of token is being generated at a given decoding step, the other heads decode the probabilities of different types of tokens. 
\begin{equation}
    o_{i}^u = softmax(W_u * h_{d_i})
\end{equation}
\begin{equation}
    o_{i}^b = softmax(W_b * h_{d_i})
\end{equation}
\begin{equation}
    o_{i}^v = softmax(W_v * h_{d_i})
\end{equation}
\begin{equation}
    o_{i}^s = softmax(W_s * h_{d_i})
\end{equation}
\begin{equation}
    t_{i} = softmax(W_t * h_{d_i})
\end{equation}
where $W_x \in R^{V_x X d_h}$ and $x : \{u,b,v,s\}$. We also treat different \textit{categories} as words in a vocabulary of size $V_c$ and therefore, $W_t \in R^{V_c X d_h}$ 

We, thus, train the model on an additional \textit{auxiliary} task of predicting the type of the token being generated at each step. Hence, the overall cross-entropy objective to decode the correct type at all steps becomes
\begin{equation}
    L_{aux} (\theta, \phi) = -\frac{1}{n}\sum_{0}^n{t_i^t * log(t_i))}
\end{equation}
where $t_i^t$ is the target type (from ground truth) of token to be predicted at this step and the probability of generating token $y_i$ is given by $o_i^{x[t_i]}$. $\phi$ refers to additional decoder parameters introduced in the model. Thus, our overall objective is now a sum of both cross entropy and auxiliary objective:
\begin{equation}
    L_{sep}(\theta,\phi) = L_{CE}(\theta) + L_{aux}(\theta,\phi)
\end{equation}

\subsection{Decoder Self Attention}
One of the key challenges for the model is to identify the relationship between the \textit{variables} it generates. A variable $A$ that is an argument in a binary predicate should be aligned with the same variable used as an argument in a unary predicate previously. One of the ways to achieve such alignment is through decoder self-attention which is an extension of the regular encoder-decoder attention. In this case, \textit{queries}, \textit{keys} and \textit{values} - all are decoder hidden states $H_d$. Therefore, we determine decoder context vector $c_d$ along with the encoder-context vector. However, one of the key differences between encoder-decoder and decoder self-attention is that while encoder-decoder attention can be applied to the whole input, decoder self-attention can only be applied on the hidden states which have been decoded so far. Just like encoder-context, decoder context is calculated by taking a weighted sum of decoder hidden states

\begin{equation}
    \beta_{ij} = \frac{e^{h_{d_i}^Th_{d_j}}}{\sum_{k=0}^{m}e^{h_{d_i}^T h_{d_k}}}
\end{equation}
\begin{equation}
        c_{d_i} = \sum_{j=0}^{i-1}\beta_{ij} * h_{d_j}
\end{equation}

The linear head on the decoder now uses both encoder and decoder contexts along with decoder hidden state to generate the final output
\begin{equation}
    s_i = W_o * [h_{d_i};c_{e_i};c_{d_i}]
\end{equation}
where $W_o \in R^{d_h + 2*d_c}$
\subsubsection{Alignment Mechanism} \label{align-mech}
Through decoder self attention, the model does not receive any explicit signal on alignment and relies only on cross-entropy objective to identify such relations between different \textit{variables}. In order to provide an explicit signal to the model, we introduce \textbf{Alignment Mechanism}. 

At each \textit{variable} decoding step, along with decoding the type of the token \textit{i.e.} variable, a linear classifier makes the decision whether this variable is aligned with any previously decoded token/variable or is an entirely new variable being generated at this step
\begin{equation}
    A_{d_i} = sigmoid(W_{ali} * h_{d_i})
\end{equation}
where $W_{ali} \in R^{1Xd_h}$. Depending on this decision by the classifier, an alignment mechanism similar to decoder self-attention performs the relational mapping between a previously decoded variable and the variable which is currently being decoded. This mapping is performed only for the \textit{variables} and not for any other category. All the previously generated hidden states of the decoder are linearly projected into a different space before calculating the position of the token with which the variable is aligned. The projection is performed to reduce the interference with encoder-decoder attention due to alignment mechanism training.
\begin{equation}
    h_{d_i}^{a} = W_{proj}*h_{d_i}
\end{equation}

The probabilities of whether a particular step $j$ aligns with the current decoding step $i$ is calculated with an attention-like formulation. 
\begin{equation}
    \gamma_{ij} = \frac{e^{h_{d_i}^{a T} h_{d_j}^a}}{\sum_{k=0}^{m}e^{h_{d_i}^{a T} h_{d_k}^a}}
\end{equation}
\begin{equation}
    c_{a_i} = \sum_{j=0}^{i-1}\gamma_{j} * h_{d_j}^a
\label{weight-indices}
\end{equation}

For every other category of tokens but \textit{variables}, the decoder heads remain the same. However, for \textit{variables}, we first calculate aligned hidden state value as
\begin{equation}
    h_{d_i}^{ali} = A_{d_i} * c_{a_i} + (1-A_{d_i}) * h_{d_i}
\label{weight-decision}
\end{equation}
 The output is, then, calculated as
 \begin{equation}
     o_v^i = W_o * [h_{d_i}^{ali};c_{e_i}]
 \end{equation}
In order to provide explicit signal during training, we train $\gamma$ and $A_{d_i}$ on the target alignment positions and decisions with a cross entropy objective
\begin{multline}
    L_{dec}(\theta,\phi,\zeta) = 
    -\frac{1}{n}\sum_{i=0}^{n}(A_{d_i}^t*log(A_{d_i}) + \\(1-A_{d_i}^t)*log(1-A_{d_i}))
\end{multline}

\begin{equation}
    L_{pos}(\theta,\phi,\zeta)  = -\frac{1}{n}\sum_{i=0}^{n}\sum_{j=0}^{i-1}A_{d_i}^t*\gamma_{ij}^t*log(\gamma_{ij})
\end{equation}
where $A_{d_i}^t$ and $\gamma_{ij}^t$ refer to ground truth decision and alignment position values and $\zeta$ refers to additional parameters introduced due to alignment mechanism.
Therefore our overall loss becomes
\begin{equation}
    L_{align}(\theta,\phi,\zeta) = L_{sep} + L_{dec} + L_{pos} 
\end{equation}

\section{Experiments}

\subsection{Dataset}
We collated a subset of SNLI \cite{bowman2015large} corpus by extracting sentences from both \textit{premise} and \textit{hypothesis} for a \textit{limited} number of examples. Eliminating duplicates, we \textit{prepared} (Refer Section \ref{section:background}) two versions of the dataset - \textbf{Small} and \textbf{Large} to examine if the proposed improvements remain consistent even on small data. In the smaller version, we prepared 138,346 instances while in the larger one, we prepared 255,501 instances for training. We used the development and test sets of SNLI \textit{as provided} but eliminated the duplicates resulting in evaluation set having 10,691 instances and test set having 10,633 instances. \\

\subsection{Implementation}
We used Pytorch\footnote{https://pytorch.org} library for implementing an auto-differentiable graph of our computations. All the models were trained with an Adam Optimizer\cite{kingma2014adam} initialized with a learning rate of $0.001$ with a decay rate of $10^{-4}$. We use an embedding size $D$ = $100$ for encoder as well as decoder embeddings in the baseline model. In our separated heads model, $D$ remained the same for encoder embeddings. However, on the decoder side, Unary and Binary predicates have an embedding size of 100 each while variable and type
embeddings are one-hot having the number of dimensions equal to their respective vocabulary sizes. Scoped entities, being very less in number, were encoded with an embedding size of 50. Our final input embedding is a concatenation of Unary, Binary, Variable, Scope and type embeddings. All dense embeddings are randomly initialized and trained from scratch.\footnote{We experimented with GloVe embeddings but it did not give further improvements} We used $d_h= d_c = 400$, $m=100$ and $n=30$.

\subsection{Results and Discussion}


\subsubsection{Evaluation Framework}
We evaluate different models through estimating the accuracy of complete match between gold standard FOL and predicted output. Due to the complex nature of the task, it is less likely that the model generates exactly the same FOL. To mitigate this, we propose to evaluate the degree of partial match between two FOLs following the intuition behind D-match and Smatch \cite{cai2013smatch}, which are widely used to evaluate AMR graphs and DRGs. We align two FOLs in bottom up manner beginning with variables. For aligning two variables, it is required that the corresponding predicates' name (in which they appear as arguments) and argument positions match. Subsequently, while aligning two predicates, we check if their arguments are aligned and their names are same. We continue to follow the same process where we align nested scope symbols (``not(" etc.). In particular, given an expected scoped entity, we determine the predicted scoped entity having maximum alignment with it based on the count of other aligned predicates and scoped entities that are contained inside them. Given an FOL, we decompose it into related pairs of the form $(n_1,n_2)$ such that $n_2$ appears inside the scope of $n_1$. For instance, a variable that is an argument in a predicate or a predicate appearing inside a scoped entity. Consequently, we estimate the number of pairs in expected FOL that can be matched with pairs in predicted FOL based on the constraint that corresponding entities in the pairs should be aligned. We select the alignment with maximum matches and report metrics (precision-recall and F1 over pair-matching) as evaluation criteria along with overall FOL accuracy.

\subsubsection{Comparison with Baseline and Ablation Studies}

We conduct a range of experiments and evaluations on different models. We show our results on both \textit{development} and \textit{test} sets in Table \ref{result-table-dev}, \ref{result-table-test}, \ref{result-table-dev-150}, and \ref{result-table-test-150}. Our \textbf{Vanilla (Baseline)} model consists of a biLSTM Encoder and a plain LSTM decoder as described in Section \ref{decoder} coupled with an encoder-decoder attention mechanism. Performing disentanglement, our \textbf{Separate Heads} model uses different linear heads on the top of LSTM decoder for different category of tokens as discussed in Section \ref{sep-heads}. Our final proposed model \textbf{Separate Heads + Align} uses our alignment mechanism on the top of \textbf{Separate Heads} and utilises the disentangled variable prediction mechanism coupled with an alignment mechanism to effectively identify the relationships between variables in binary predicates and their unary counterparts. We also conduct ablations on the \textbf{Vanilla} and \textbf{Separate Heads} models by incrementally adding both decoder self-attention and alignment mechanisms.

Evidently, our final model \textbf{Separate Heads + Align} convincingly outperforms all described models and improves the baseline by $\sim8$ F-1 points. Decoder self-attention, even though, improves \textbf{Vanilla} Model does not provide any improvements when used with \textbf{Separate Heads}. This can be attributed to its inability to incorporate decoder level information which probably becomes factorized automatically during training through using separate heads. However, it provides improvements over \textbf{Vanilla} by a good margin but still only matches or remains inferior to the standalone \textbf{Separate Heads} model. \textbf{Align Mechanism} manages to provide a huge boost to the \textbf{Separate Heads} model by improving it by $\sim 5$ F-1 points. However, performance deteriorates when used with \textbf{Vanilla} model since its ability to align variables \textit{only} vanishes in this setup which we find critical for its working. We further note that by increasing the size of training data, the performance increases uniformly with our final model achieving the best F-1 of $\sim 73$ and an overall accuracy of $\sim 63\%$.


\setlength\tabcolsep{4.5pt}
\begin{table}[hbt!]
\centering
\resizebox{225pt}{!}{%
\begin{tabular}{lllll}
\hline \textbf{Model} & \textbf{Precision} & \textbf{Recall} & \textbf{F-1} & \textbf{Accuracy}\\
\hline
\textbf{Vanilla (Baseline)}& 65.48 & 65.15 & 65.31 & 59.26\\
+ Self Attention & 67.86 & 66.70 & 67.28 & 60.74\\
+ Align Mechanism & 62.13 & 61.06 & 61.59 & 56.98\\
\hdashline
\textbf{Separate Heads} & 68.48 & 66.81 & 67.64 & 60.77\\
+ Self Attention & 67.11 & 65.87 & 66.48 & 60.02\\
\hdashline
\textbf{Separate Heads + Align} & \textbf{73.68} & \textbf{72.17} & \textbf{72.92} & \textbf{63.26}\\
\hline
\end{tabular}%
}
\caption{\label{result-table-dev} Results showing overall accuracy and F1-Scores of different models (trained on \textbf{Large} dataset) on development dataset }
\end{table}
\setlength\tabcolsep{4.5pt}
\begin{table}[hbt!]
\centering
\resizebox{225pt}{!}{%
\begin{tabular}{lllll}
\hline \textbf{Model} & \textbf{Precision} & \textbf{Recall} & \textbf{F-1} & \textbf{Accuracy}\\
\hline
\textbf{Vanilla (Baseline)}& 66.43 & 65.84 & 66.13 & 60.45\\
 + Self Attention & 68.56 & 67.23 & 67.89 & 61.18\\
 + Align Mechanism & 61.74 & 60.60 & 61.17 & 56.79\\
 \hdashline
\textbf{Separate Heads} & 69.18 & 67.63 & 68.40 & 61.14\\
 + Self Attention & 66.92 & 65.69 & 66.30 & 60.12\\
 \hdashline
\textbf{Separate Heads + Align} & \textbf{74.05} & \textbf{72.53} & \textbf{73.28} & \textbf{63.24}\\
\hline
\end{tabular}%
}
\caption{\label{result-table-test} Results showing overall accuracy and F1-Scores of different models (trained on \textbf{Large} dataset) on Test dataset }
\end{table}

\setlength\tabcolsep{4.5pt}
\begin{table}[hbt!]
\centering
\resizebox{225pt}{!}{%
\begin{tabular}{lllll}
\hline \textbf{Model} & \textbf{Precision} & \textbf{Recall} & \textbf{F-1} & \textbf{Accuracy}\\
\hline
\textbf{Vanilla (Baseline)}& 59.54 & 58.14 & 58.83 & 52.99\\
\textbf{Separate Heads} & 62.00 & 60.58 & 61.28 & 55.20\\
\textbf{Separate Heads + Align} & \textbf{65.75} & \textbf{64.36} &\textbf{65.05} &\textbf{55.30} \\
\hline
\end{tabular}%
}
\caption{\label{result-table-dev-150} Results showing overall accuracy and F1-Scores of different models (trained on \textbf{Small} Dataset) on development dataset }
\end{table}

\setlength\tabcolsep{4.5pt}
\begin{table}[hbt!]
\centering
\resizebox{225pt}{!}{%
\begin{tabular}{lllll}
\hline \textbf{Model} & \textbf{Precision} & \textbf{Recall} & \textbf{F-1} & \textbf{Accuracy}\\
\hline
\textbf{Vanilla (Baseline)}& 59.87 & 58.41 & 59.13 & 52.92\\
\textbf{Separate Heads} & 62.56 & 61.10 & 61.82 & 55.94\\
\textbf{Separate Heads + Align} &\textbf{66.45} &\textbf{65.05} & \textbf{65.74}&\textbf{56.10} \\
\hline
\end{tabular}%
}
\caption{\label{result-table-test-150} Results showing overall accuracy and F1-Scores of different models (trained on \textbf{Small} Dataset on test dataset }
\end{table}


\subsubsection{Analysis}
We perform additional experiments to analyse the results observed. 
We conduct two sets of analysis - Variation of F-1 score with input length and \textit{Perturbed} training to establish the robustness of our proposed method.
\begin{figure}[htbt!]
    \centering
    \includegraphics[width=1.1\linewidth]{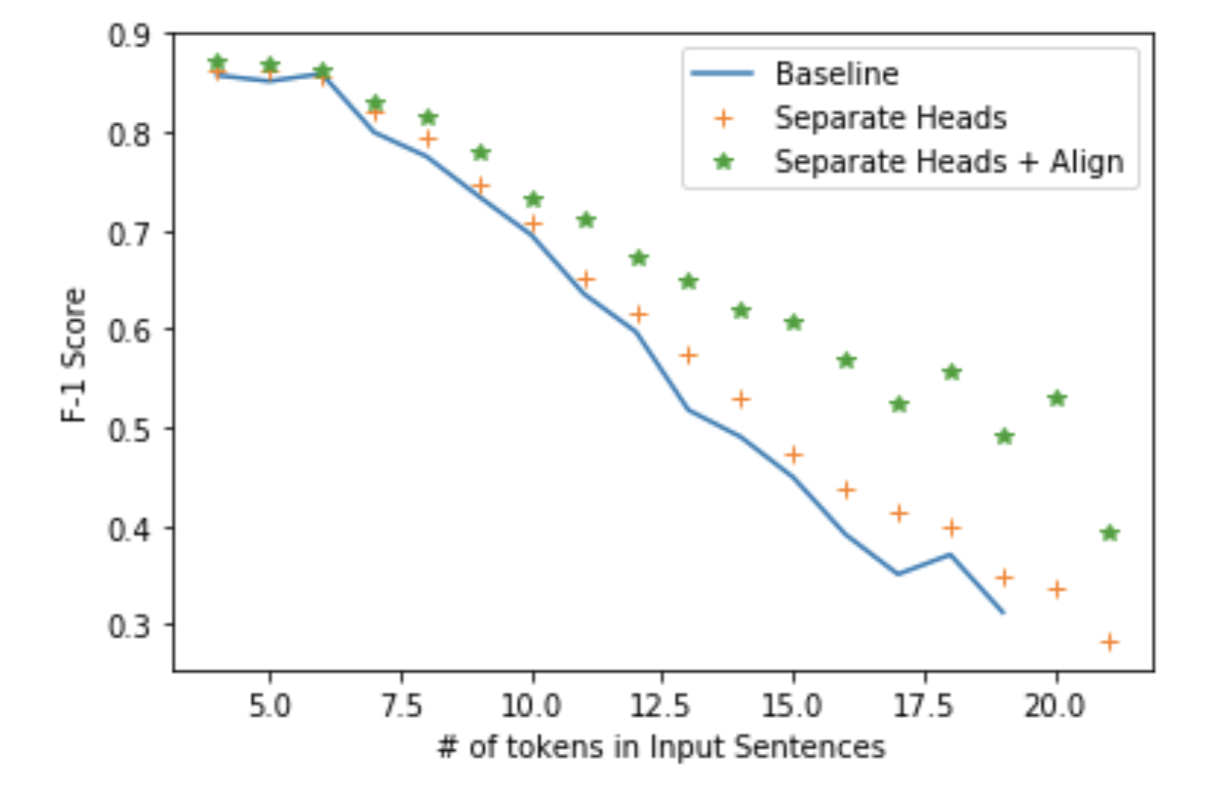}
    \caption{Variation of output F-1 Score with input length on Test Dataset}
    \label{fig:plot}
\end{figure}

\noindent\textbf{Variation with Input Length:} Evidently, our proposed models are relatively much more robust to increase in length in the input sentence as shown in Fig. \ref{fig:plot}. This can be attributed to many factors - increased model capacities as well as their abilities to process different categories of output tokens separately giving better long range dependencies and less confusion in generating many variables over FOL owing to better alignment across the sequence.
\setlength\tabcolsep{4.5pt}
\begin{table}[hbt!]
\centering
\resizebox{225pt}{!}{%
\begin{tabular}{lllll}
\hline \textbf{Model} & \textbf{Precision} & \textbf{Recall} & \textbf{F-1} & \textbf{Accuracy}\\
\hline
\textbf{Vanilla(Baseline)} & 63.75 & 62.32 & 63.03 & 57.95\\
\hdashline
\textbf{Separate Heads} & 67.39 & 65.59 & 66.48 & 61.44\\
\textbf{Separate Heads + Align} & 72.90 & 71.95& 72.42 & 63.14 \\
\hline

\end{tabular}%
}
\caption{Results on Test set showing both accuracy and F1-Scores with perturbed training}
\label{results-adversary}
\end{table}

\noindent\textbf{Perturbed Training:} It has been noticed in literature (\cite{jia2017adversarial, niven2019probing} that neural models sometimes exploit trivial patterns in outputs/inputs to fool and provide pseudo-improved results. One such pattern could be presence of variables like \textit{A} and \textit{B} with some specific unary and binary predicates. In order to disturb such patterns, we randomly permute the presence of such variables in the ground truth during training. Our \textbf{baseline} model indeed shows a significant drop in results (Table \ref{results-adversary}). On the other hand, our other two main models do not show such large drop proving their robustness to such disturbances.

\section{Related Work}
Early semantic parsers were majorly rule based \cite{johnson1984natural,woods1973progress,thompson1969rel} using grammar systems \cite{waltz1978english,hendrix1978developing}, employing shallow pattern matching \cite{johnson1984natural} and parse tree to generate database query language \cite{woods1973progress}. These were succeeded by data driven learning techniques which use language data paired with meaning representations and can be broadly classified into statistical methods \cite{thompson2003acquiring,zettlemoyer2012learning,zelle1996learning,kwiatkowski2010inducing} and neural approaches \cite{kovcisky2016semantic,dong2016language,jia2016data,buys2017robust,cheng2017learning,liu2018discourse,li2018seq2seq}. \cite{zettlemoyer2012learning} proposed to use sentences and their lambda calculus expressions to learn a log linear model through probabilistic CCG to rank different parses for a given sentence using simple features such as count of lexical entries. Additionally, captioned videos have been used to perform visually grounded semantic parsing \cite{ross2018grounding}. Feedback based semantic parsing has been done to facilitate continuous improvement in the quality of parse through conversations \cite{artzi2011bootstrapping} and user interaction \cite{iyer2017learning,lawrence2018improving}.

Neural approaches alleviate the need for manually defining lexicons and can further be categorized based on the structure of parse into sequential parse prediction \cite{jia2016data,kovcisky2016semantic} and graph structure decoding which tailor network architecture to utilize the syntactic structure of meaning representation. \cite{yin2017syntactic,alvarez2016tree,rabinovich2017found}. \cite{dong2016language} proposed SEQ2TREE to generate domain-specific hierarchical logical form by introducing parenthesis token and parent connections to recursively generate sub-trees. \cite{rabinovich2017found} introduced a dynamic decoder whose components are composed depending on generated tree parse. \cite{liu2018discourse} parse DRSs using dedicated hierarchical decoders to generate partial structure first before the semantic content. We instead make the model disambiguate syntactic types (unary and binary predicates, variables, scope symbols) through performing category type prediction as an auxiliary task and using separate prediction heads. Constrained decoding using target language syntax and grammar rules has been explored \cite{yin2017syntactic,xiao2016sequence}. Copy-mechanism \cite{gu2016incorporating} has been used to facilitate the generation of out of vocabulary entities through encoder attention \cite{jia2016data}. However, our variable alignment mechanism is different since it constrains the model to align binary predicate arguments with previously generated unary structures (alignment happening at decoder level) through specifying explicit loss on whether to align and where to align.

\section{Conclusion}
In this work, we examined the capability of neural models on the task of parsing First-Order Logic from natural language sentences. We proposed to disentangle the representations of different token categories while generating FOL output and used category prediction as an auxiliary task. We utilized token factorization to build an alignment mechanism which effectively manages to capture the relationship between variables across different predicates in FOL. Our analysis showed the difficulties faced by neural networks in modeling FOL and ways to tackle them. We also experimented by introducing a perturbation in inputs in order to examine the robustness of different proposed models. In a bid to promote research further in the area, we aim to release our code as well as data publicly. 



\bibliography{anthology,acl2020}
\bibliographystyle{acl_natbib}


\end{document}